\documentclass[letterpaper, 10 pt, conference]{ieeeconf}
\IEEEoverridecommandlockouts    % This command is only needed if you want to use the \thanks command
\usepackage{color}
\usepackage{graphics} % for pdf, bitmapped graphics files
\usepackage{epsfig} % for postscript graphics files
\usepackage{mathptmx} % assumes new font selection scheme installed
\usepackage{times} % assumes new font selection scheme installed
\usepackage{amsmath} % assumes amsmath package installed
\usepackage{amssymb}  % assumes amsmath package installed
\usepackage{graphicx}
\usepackage{multirow, makecell}
\usepackage{array}
\usepackage{booktabs} % For Table
\usepackage{comment}
\usepackage[linesnumbered,ruled]{algorithm2e} % For pseudocode and algorithms
\newcolumntype{M}[1]{>{\centering\arraybackslash}m{#1}}
\usepackage{caption}
\usepackage{subcaption}
\usepackage{hyperref}

\usepackage{tikz}
\usetikzlibrary{shapes,arrows,calc}

\title{\LARGE \bf
A Data-driven, Falsification-based Model of Human Driver Behavior}

\author{Nauman Sohani$^{*}$, Geunseob (GS) Oh$^{*}$, Xinpeng Wang
\thanks{Nauman Sohani, Geunseob (GS) Oh, and Xinpeng Wang are with the Department of Mechanical Engineering, University of Michigan, Ann Arbor, MI 48109 USA, Email: \{ \href{nsohani@umich.edu}{nsohani}, \href{gsoh@umich.edu}{gsoh}, \href{xinpengw@umich.edu}{xinpengw}\}@umich.edu}
\thanks{$^{*}$The authors contributed equally to this work.}
}

\begin{document}

\maketitle

\begin{abstract}
We propose a novel framework to differentiate between vehicle trajectories originating from human and non-human drivers by constructing a data-driven boundary using parametric signal temporal logic (STL). Such construction allows us to evaluate the trajectories, detect rare-events, and reduce the uncertainty of driver behaviors when it assumes the form of a disturbance in control synthesis and evaluation problems. We train a classifier that separates admissible (i.e. human) examples - which arise from real-world demonstrations - and inadmissible (i.e. non-human) examples that are generated by falsifying specifications synthesized from the same real-world driving data. Proceeding in this fashion allows for finding a reasonable boundary of human behaviors exhibited in real-world driving records. The framework is demonstrated using a case study involving a human-driven vehicle approaching a signalized intersection.
\end{abstract}

\section{Introduction} \label{introduction}
The field of human driver research has received significant attention, in part due to its relevance to connected and automated vehicles (CAVs) and subsequent problems of path-planning and control synthesis. Consequently, there is a significant body of research in the field of modeling human driver behavior that has leveraged different techniques, such as dynamic system modeling \cite{macadam2003}; neural networks \cite{khodayari2012}, \cite{oh2019}; stochastic processes \cite{pentland1999}; and inverse reinforcement learning \cite{sadigh2016}. Much of the prior work in the field has focused on predicting likely actions based on inference from driver studies or real-world observations of human drivers. Unfortunately, ``interesting" edge cases are rare events and may not be explicitly captured or reproduced in the aforementioned approaches. Therefore, we advocate a mapping as in \cite{chen2018}, which leverages real-world driving data to construct a realistic set of trajectories which accommodate the reactive and uncertain nature of human drivers. Such a method can be extended to evaluating controllers by sampling rare, (and likely dangerous) events.

In contrast to the differential game setting of \cite{chen2018}, we generate examples of non-human behaviors using falsification. The literature on cyberphysical system verification is substantial, and several mature toolboxes have been developed to address the falsification problem \cite{annapureddy2011}, \cite{breach}. Furthermore, recent literature has addressed the synthesis of precise specifications by searching over parameters for which template formulae are falsified \cite{hoxha2017}, or satisfied \cite{jin2015} by a given system. Herein, parameter synthesis is used to precisely describe human driver behavior by studying real-world examples; then falsifying these specifications generates possible non-human actions. The observed and generated examples are subsequently used in the construction of a classifier.

Such an approach has consequences for control synthesis and evaluation. Given a state-dependent description of human driver behavior, we can compute sets of interest, such as initial conditions from which a human disturbance can initiate collisions. These scenarios are instructive to test the robustness of a path-planner or a controller. There are philosophical similarities between this strategy and the work developed in \cite{chou2018}.

Motivated by \cite{oh2019} and \cite{oh2018}, we demonstrate the proposed framework on a case study of a human-driven vehicle (HV) approaching a signalized intersection. In this setting, the leading HV plays the role of a disturbance signal to the following controlled vehicle, which desires a safe, fuel-optimal policy. Our objective is to obtain a set-valued, state-dependent mapping that describes human actions using the aforementioned classification approach; such a mapping can be conceived as a driver model, which can be utilized for synthesizing a fuel-optimal safe controller as in \cite{oh2018}. In the absence of such a mapping, we may resort to a worst-case approach based only on the physical limitations of a given situation or one of the aforementioned probabilistic methods.

The remainder of this paper is structured in the following way: Section \ref{problem_formulation} gives an overview of the problem under study. Section \ref{mathematical_preliminaries} summarizes the key mathematical tools that are leveraged to solve the problem. Section \ref{solution_approach} consolidates the methods of Section \ref{mathematical_preliminaries} to detail the solution approach alluded to in Section \ref{problem_formulation}. Section \ref{results} discusses results and practical considerations of our solution approach. Section \ref{conclusion} offers concluding remarks and plans for future work.

\section{Problem Formulation}\label{problem_formulation}
The objective of this work is to systematically determine a set-valued, state-dependent bound on human driving behavior. We consider the car-following problem in the vicinity of a signalized intersection, as in \cite{oh2018}. A robust control policy in this setting is concerned with implementing a fuel-optimal policy while respecting possible acceleration actions of the leading vehicle, i.e. the disturbance.

We reason that a driver is unlikely to modify his/her behavior based on the actions of a trailing vehicle, but will be affected by other factors such as the state of a traffic light or length of the vehicle queue already formed at an intersection \cite{oh2019}. We make the following assumptions on the motion of the HV: (1) The HV passes through the intersection, i.e. no left/right turning actions; (2) The HV does not change lanes; (3) The acceleration of the HV is determined only by a short history of the state of the traffic light and its current kinematic state.

Formally, the problem can be stated as the construction of a mapping from a sequence of states to an admissible subset of the acceleration input of HV at the next time instance:
\[
f: {\displaystyle \prod_{t=t_0-h}^{t=t_0} X\xrightarrow{} 2^U}.
\]

\subsection{System Model}\label{system_model}
\begin{figure}[ht]
  \centering
      \includegraphics[width=0.90\linewidth]{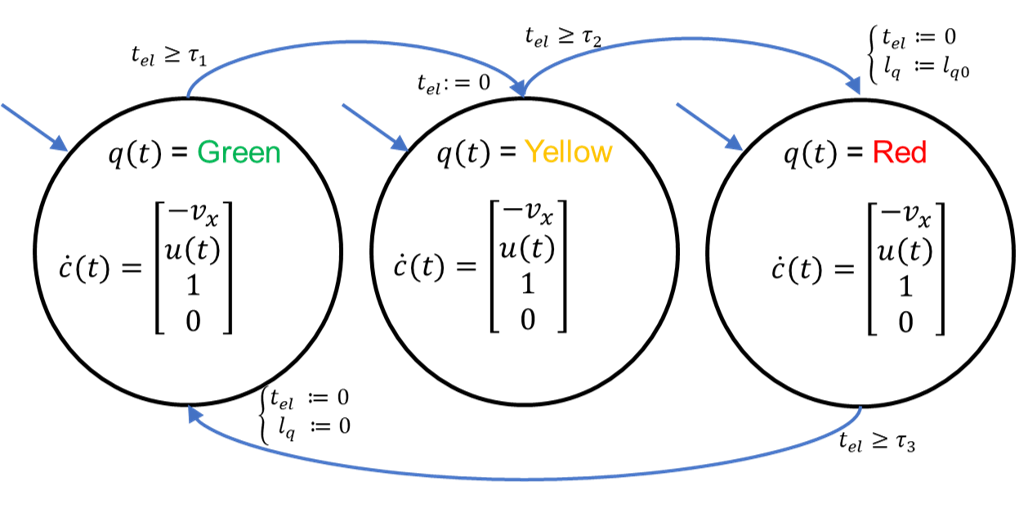}
  \caption{Hybrid automaton representation of $\Sigma$}{\label{fig:syst}}
\end{figure}

The system, $\Sigma$, capturing the evolution of the states affecting a human driver can be represented as a hybrid automaton (Figure \ref{fig:syst}). Relevant quantities of $\Sigma$ are summarized in Table \ref{tab:sigma}. In this study, the signal light will cycle through the different colors on a fixed schedule reflecting the most frequent values of signal phasing and timing data from the Safety Pilot Model Deployment (SPMD) database (c.f. Section \ref{SPMD}). Among the four continuous states, the estimated length of the queue formed at the intersection, $l_q$, is 0 during green and yellow lights, and a constant value on red lights. Note that this model is agnostic to the specific vehicle model: instead, $\Sigma$  models the evolution of states that represent the driver's surroundings and impact the HV's acceleration.

\begin{table}
    \begin{tabular}{ c | c | c | c }
    Qty. & Description & Type & Range \\ \hline
    $d_x$ & Distance to intersection & Continuous state & [0, $\overline{d}]$ \\ \hline
    
    $v_x$ & Velocity & Continuous state & [0, $\overline{v}]$ \\ \hline
    
    $t_{el}$ & Time since last $s_{TL}$ change & Continuous state & [0, $\infty)$ \\ \hline
    
    $l_q$ & Traffic queue at intersection & Continuous state & [0, $d_x]$ \\ \hline
    
    $s_{TL}$ & State of traffic light & Discrete state & $\{G,Y,R\}$ \\ \hline
    
    $u(t)$ & Acceleration & Input & [$\underline{a},\overline{a}$] \\ \hline
    
    \end{tabular}
    \caption{Summary of important quantities of $\Sigma$}
    \label{tab:sigma}
\end{table}

\subsection{Overview of Real-World Driving Data}\label{SPMD}

The human driving data were collected from the Safety Pilot Model Deployment (SPMD), a large-scale connected vehicle study conducted in the Ann Arbor, MI area \cite{Bezzina2014}. For this work, 556 eastbound trajectories from three weeks in 2014 were extracted from instrumented vehicles and synchronized with V2X communication units installed at the Fuller-Bonisteel intersection (map available at: \url{https://www.google.com/maps/@42.2873631,-83.7196829,19z}). Note that the extracted data were selected only on the premises of data integrity, i.e. no further screening criteria were applied.

\subsection{Overview of Proposed Method}
The problem of constructing a state-dependent set of HV acceleration inputs is framed as finding a boundary between human and non-human driving behavior, and subsequently translated into one of classification. SPMD provides examples of human driving traces, i.e. positive examples for the classification; on the other hand, generating negative training examples for the classification problem, i.e. the driving traces that are ``non-human", is less straightforward.

The fundamental assumption of our framework is that HVs satisfy certain specifications representing driving norms, etc. We attempt to capture these specifications using a set of Parametric Signal Temporal Logic (PSTL) formulae. A feasible parameter set for these PSTL formulae is synthesized from analysis of real-world (naturalistic) driving data; the boundary of the parameter set is used to convert PSTL formulae to Signal Temporal Logic (STL) formulae. Then falsifications of STL formulae represent violations of traffic norms that humans are assumed to satisfy. Consequently, such violations constitute negative training examples for the classifier. The mapping to human driver actions then corresponds to those actions for which the state-action tuple is classified as ``human" behavior.

\section{Mathematical Preliminaries} \label{mathematical_preliminaries}
Central to this approach is the construction of precise specifications that represent human driver behavior and subsequent classification of state-action tuples as ``human" or ``non-human". In the following, we give a brief overview of the mathematical tools employed in this framework. More details can be found in \cite{asarin2011} and \cite{fages2009}. The interplay between the tools used to achieve our objective will be described in more detail in Section \ref{solution_approach}.

\subsection{Parametric Signal Temporal Logic}
STL specifications can be conceived as constraints on a signal, $x(t):\mathbb{R}_+ \rightarrow \mathbb{R}^n$, as it evolves in time \cite{donze2014}. Such a constraint can be captured by inequalities, $\mu$, of the form $\mu := f(x(t)) \geq \pi$, where $\pi \in \mathbb{R}$. The syntax for building specifications can be defined inductively as:
\[
\phi := \top ~|~ \mu_{\pi} ~|~ \lnot \phi ~|~ \phi \wedge \psi ~|~ \phi \mathbf{U}_{[\tau_1,\tau_2]}\psi
\]
where the subscript of $\mu_{\pi}$ is used to emphasize the dependence of the constraint on the parameter $\pi$. The main distinction between STL formulae and PSTL formulae is that in the latter, some of the parameters, such as the scale parameter $\pi$, and time parameters $\tau_1$ and $\tau_2$, are left unspecified. The semantics of (P)STL formulae are given as:
\[
\begin{array}{lcl}
x(t) \models \mu_{\pi}	&	\Leftrightarrow & f(x(t)) \geq \pi \\
x(t) \models \lnot \mu_{\pi} & \Leftrightarrow & f(x(t)) < \pi \\
x(t) \models \phi \wedge \psi & \Leftrightarrow & x(t) \models \phi \text{ and } x \models \psi \\
x(t) \models \phi \mathbf{U}_{[\tau_1,\tau_2]}\psi & \Leftrightarrow & \exists t'\in [t+\tau_1, t+\tau_2]: x(t') \models \psi \\ & & \text{ and } \forall t'' \in [t,t']: x(t) \models \phi.
\end{array}
\]
We also consider the operators \emph{always} and \emph{eventually}:
\[
\begin{array}{rcl}
\lozenge_{[\tau_1,\tau_2]}\phi & := & \top \mathbf{U}_{[\tau_1,\tau_2]}\phi\\
\square_{[\tau_1,\tau_2]}\phi & := & \lnot(\lozenge_{[\tau_1,\tau_2]}\lnot(\phi)).
\end{array}
\]

(P)STL introduces a robustness metric $\rho(\phi,x(t))$, to further refine Boolean satisfaction of a specification. This is accomplished by the following \emph{quantitative semantics}:
\[
\begin{array}{rcl}
\rho(\mu_{\pi},x(t)) & = & f(x(t))-\pi \\
\rho(\lnot \phi, x(t)) & = & -\rho(\phi,x(t))\\
\rho(\phi,\psi,x(t)) & = & \min(\rho(\phi,x(t)),\rho(\psi,x(t))) \\
\rho(\phi \mathbf{U}_{[\tau_1,\tau_2]}\psi,x(t)) & = & \sup_{t' \in [t+\tau_1,t+\tau_2]}(\min(\rho(\psi,x(t)),\\ & & \inf_{t''\in [t,t']}(\rho(\phi,x(t)))) \\
\end{array}
\]
The positive (or negative) sense of $\rho(\phi,x(t))$ captures Boolean satisfaction (or violation) of the specification, and the absolute value captures the robustness with which the signal satisfies (or violates) the specification.

\subsection{Parameter Synthesis}
The parameter synthesis problem is one of finding the set of parameters which result in \emph{tight} satisfaction of a PSTL specification by signals. In particular, we consider specifications $\rho(\phi,x(t))$ that monotonically increase or decrease with a specific parameter. For such problems, parameter synthesis can be reduced to a generalized binary search \cite{asarin2011}, \cite{jin2015}.  For this work, we use the methods developed and implemented in the \textsc{Breach} toolbox \cite{breach}.

\subsection{Falsification}\label{section: falsification}

The falsification problem can be thought of as dual to that of parameter synthesis. The objective of this problem is to find an input signal which results in the violation of a given specification. In \cite{jin2015}, this is formulated as an optimization problem on $\rho(\phi,x(t))$ over input signals $u(t)$:
\[
\begin{array}{cl}
\text{minimize} & \rho(\phi(p),\Sigma(u(t))) \\
\text{s.t.} & u(t) \in \mathcal{U} \\
			& p \in \mathcal{P}.
\end{array}
\]
A negative $\rho^\star$ indicates a specification violation and the pair $(x(t),u(t))$ is referred to as a counterexample. In general, the falsification problem is undecidable, and the \textsc{Breach} toolbox may not find a counterexample even if one exists. 

\subsection{Classification}
The classification problem seeks to identify boundaries between distinct classes given labeled examples of valid class members. In this work, we operate on time-series of the state and control input, $X(t)=[x(t), u(t)]$, and check for membership in the aforementioned classes.

\begin{figure}[ht]
	\centering
	\begin{subfigure}{0.48\linewidth}
		\includegraphics[width=\linewidth]{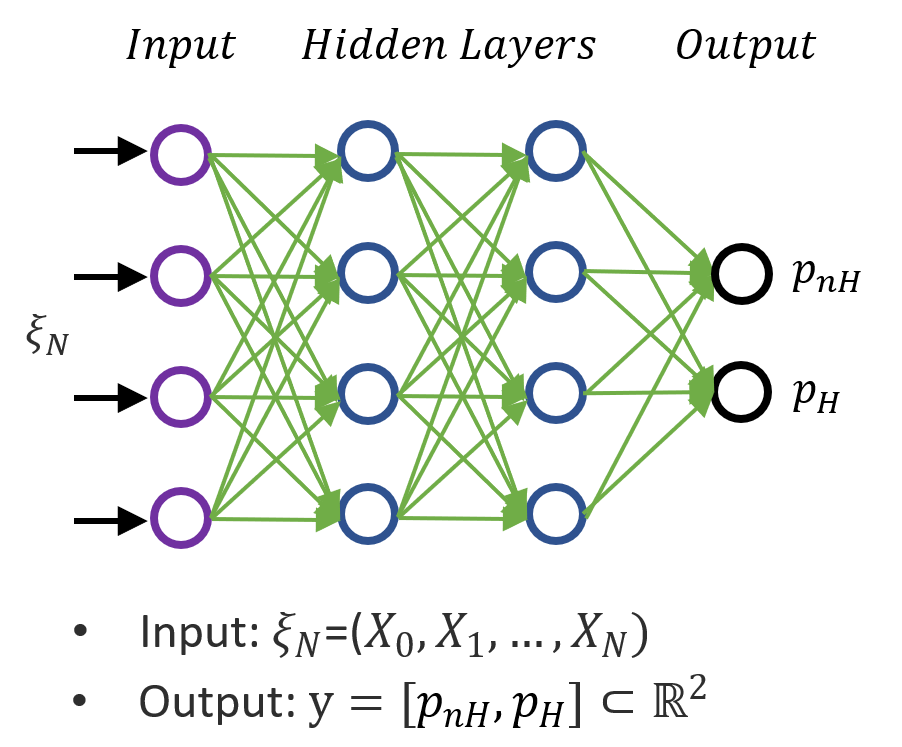}
		\label{fig:MLP}
		\caption{Feed-forward Neural Network Classifier}
	\end{subfigure}
    \begin{subfigure}{0.48\linewidth}
		\includegraphics[width=\linewidth]{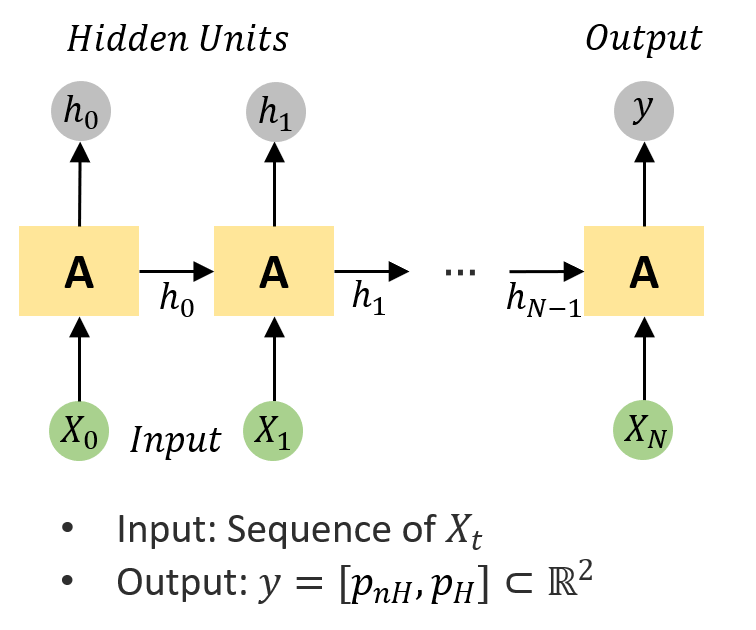}
		\label{fig:RNN}
		\caption{Recurrent Neural Network Classifier}
    \end{subfigure}
	\caption{Two classifiers are used to classify ``human" traces from ``non-human" traces}
	\label{fig:Classifier}
\end{figure}

The former can be modeled using a feed-forward neural network (or multi-layer perceptron, MLP), and the latter using a recurrent neural network (RNN) (Figure \ref{fig:Classifier}). The MLP is a generic non-linear function approximator and is widely used for regression and classification; however, it is best equipped to handle only instantaneous snapshots of a time-series. On the other hand, RNN is a class of artificial neural networks where connections between nodes form a directed graph along a sequence. This allows for incorporating the temporal dynamic behavior of a time sequence. Unlike feed-forward neural networks, RNN can use their internal hidden units to process sequences of inputs.

In this project, we demonstrate both methods of classification and compare the performance of the two classifiers. The output can be either modeled using a binary variable $y(t) \in {0, 1}$ (where 0 indicates ``non-human", and 1 indicates ``human") or using two distinct variables $y(t) = [p_{nH}, p_H]^T$ where $p_{nH}\in[0,1]$ and $p_{H}\in [0,1]$ indicate the probability of the given trace to be non-human and human, respectively. Note that $\sum(y_i(t)) = 1$. In this work, we will use the notion of the output that is most convenient in the relevant discussion.

% Define block styles
\tikzstyle{decision} = [diamond, draw, fill=blue!20, 
text width=4.5em, text badly centered, node distance=3cm, inner sep=0pt]
\tikzstyle{block} = [rectangle, draw, fill=blue!20, 
text width=8em, text centered, minimum height=4em]
\tikzstyle{line} = [draw, -latex']
\tikzstyle{cloud} = [draw, ellipse,fill=red!20, node distance=3cm,
minimum height=2em]
\tikzstyle{input} = [rectangle, draw, fill=red!20, 
text width=8em, text centered, rounded corners, minimum height=4em]	
\tikzstyle{output} = [rectangle, draw, fill=green!20, 
text width=8em, text centered, rounded corners, minimum height=4em]	

\section{Solution Approach}\label{solution_approach}
As described in Section \ref{problem_formulation}, we seek to construct a set of driver inputs given a finite history of the states and inputs, 
\begin{equation}\label{drivermodel1}
f(x(t_0-h),...,x(t_0),u(t_0-h),...,u(t_0-1)) = [\underline{u(t_0)},\overline{u(t_0)}].
\end{equation}

In our solution, we will consider 3-second intervals consisting of tuples of the state and control input. Suppose the classifier takes the form 
\begin{equation}\label{classifiermodel1}
g(x(t_0-h),...,x(t_0),u(t_0-h),...,u(t_0))=g(\bar{x},\bar{u})    
\end{equation}
and maps arguments to the real numbers. In this setting, a positive value of (\ref{classifiermodel1}) implies that the tuple is an example of human behavior and a negative value implies that the tuple is an example of non-human behavior. Hence, the boundary of human behavior is the set of tuples $(\bar{x},\bar{u})$ for which the classifier evaluates to zero:
\begin{equation}\label{classifierboundary1}
X^{\text{boundary}} = \{(\bar{x},\bar{u}) ~|~ g(\bar{x},\bar{u})=0\}
\end{equation}
(Note: this approach can be adapted for classifiers which produce an output in $[0,1]^2$, i.e. expressing the probability of an input tuple belonging to either class, by finding the set of tuples for which the output is $[0.5, 0.5]^T$.)

Given the classifier in (\ref{classifiermodel1}), the set-valued driver behavior mapping, $f$, can be defined as $f(\bar{x}) = [\underline{u}, \overline{u}]\subseteq [\underline{a},\overline{a}]$ where the lower limit, $\underline{u(t_0)}$, is found from (\ref{drivermodel1}) and (\ref{classifierboundary1}):
\[
\underline{u(t_0)} = 
\min\{u(t_0) ~|~ (\bar{x},\bar{u}) \in X^{\text{boundary}}\}
\]
In practice, we seek a compact set to represent the range of inputs as in (\ref{drivermodel1}). Moreover, from the perspective of control synthesis treating the driver as a disturbance for our case study, we are interested in the lower limit of the human acceleration because the lowest acceleration, i.e. hardest brake, would be a dangerous disturbance. Therefore, one method may be to initiate a root-finding routine for $g$ initialized at the lowest acceleration permissible by the road friction. These details are explored in more detail in Section \ref{results}.

The selection of negative training examples for classification requires actions which no human would undertake given the state. In order to generate such examples, we posit that humans generally satisfy a set of specifications; then violations of these specifications are \emph{candidates} for negative training examples. In this study, we consider linear-time properties representing traffic norms. While not every violation of a traffic norm constitutes non-human behavior, we argue that violation of traffic norms is a necessary condition for non-human behavior. (Note the distinction between traffic rule and traffic norm: the former refers to laws whose violations would result in traffic citations, whereas the latter refers to common driving practices that we sought to discover through parameter synthesis. The interplay between traffic rules and norms is discussed in more detail in Section \ref{results}). The basis for counterexample generation then is falsification of specifications that human drivers satisfy. The problem of creating precise specifications for subsequent falsification is posed as one of parameter synthesis where the template reflects some traffic norm. Note that this is a different flavor from the requirement mining methods of \cite{jin2015}: Jin et al. developed a framework to synthesize the requirements to which legacy controllers were developed for subsequent analysis (possibly using formal methods). On the contrary, in this work, the controller under study is the human driver itself, and the falsifier becomes a proxy for ``non-human" behavior.

It is reasonable to ask whether the synthesized set of specifications itself can be used to define the boundary of human driver behavior. We argue that such a method may encounter the following issues: (1) the set of specifications would have to be ``complete" in some sense, i.e. it should represent all the norms that human drivers follow; (2) we argue that violation of traffic norms is a necessary condition for identifying non-human behavior but it is not sufficient: hence there may be examples of violating behavior that is valid human behavior. In brief, the authors are not aware of methods to quantify the quality of the set of specifications but this may be possible with a classification approach as described in Section \ref{conclusion}.

\begin{figure}
	\centering
	\begin{tikzpicture}[node distance = 3cm, auto]
	% Place nodes
	\node [input] (postraces) {$X^+$: traces of human driver behavior};
	\node [block, below of=postraces] (paramsynth) {Parameter synthesis};
	\node [input, left of=paramsynth, node distance = 4 cm] (pstl) {$\Phi$: PSTL representations of traffic norms};
	\node [block, below of=paramsynth] (falsify) {Counterexample generation using falsification};
	\node [input, left of=falsify, node distance = 4 cm] (model) {$\Sigma$: dynamical model};
	\node [block, below of=falsify] (classify) {Classification};
	\node [output, below of=classify] (behaviormodel) {$f(\cdot)$: driver behavior model};
	% Place intermediate coordinate
	\coordinate (rightofpostraces) at ($(postraces.east)+(1.0 cm,0 cm)$);
	% Draw edges
	\path [line] (postraces) -- (paramsynth);
	\path [line] (pstl) -- (paramsynth);
	\path [line] (paramsynth) -- node {$\mathcal{P}(\phi),$ $\forall \phi \in \Phi$}(falsify);
	\path [line] (model) -- (falsify);
	\path [line] (falsify) -- node {$X^-$}(classify);
	\path [line] (classify) -- node {$g(\underline{x},\underline{u})$}(behaviormodel);
	\path[line] (postraces)--(rightofpostraces)|-(classify.east);
	\end{tikzpicture}
	\caption{Description of solution approach}
	\label{fig: workflow}
\end{figure}
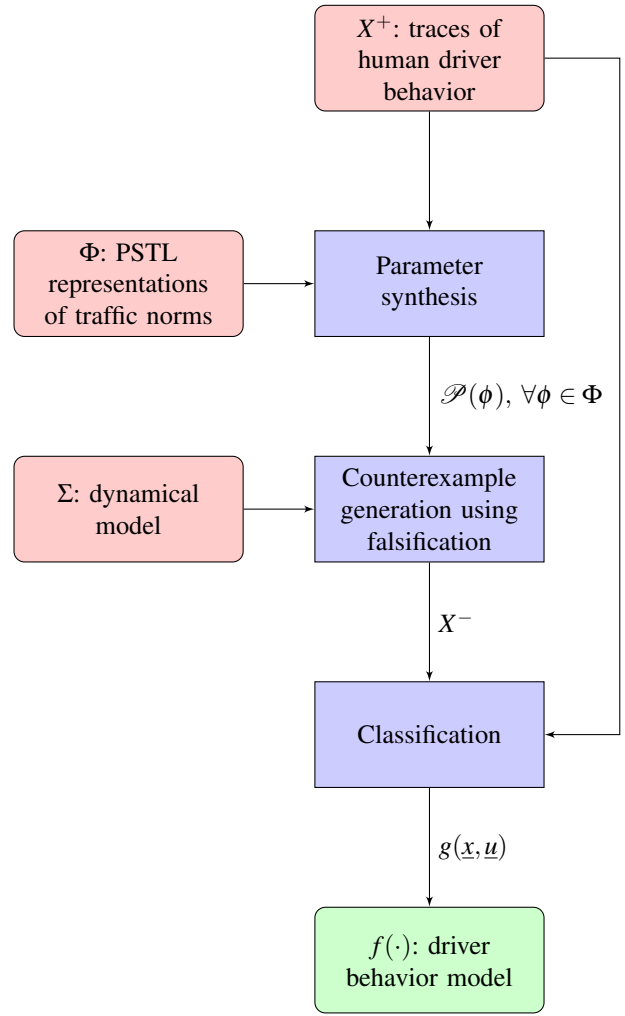

The overall work-flow is described in the context of Figure \ref{fig: workflow}. Red boxes represent inputs; these are traces of human driver behavior, specification templates representing driver behavior in the form of PSTL formulae, and a dynamical model in Simulink with an interface to \textsc{Breach}. The traces and PSTL formulae are inputs to the parameter synthesis problem. The output of this block is a set of feasible parameters for a given specification. The feasible parameter set and specifications are considered in the falsification problem wherein we seek a control signal to violate the specification; consequently the falsifier is a proxy for a non-human driver. These negative training examples are combined with the positive training examples used for parameter synthesis in the classification block wherein we seek the aforementioned function $g$. Finally $f$ is obtained through $g$ using the querying process described above. The first two blue blocks corresponding to parameter synthesis and falsification are treated separately from the last blue block corresponding to classification. Some iteration between the two, i.e. sampling traces belonging to the ``human driver" class and including these in $X^+$ for subsequent classifier construction, may be considered in future work - this is discussed in more detail in Section \ref{conclusion}. A counterexample generation strategy corresponding to the first two blue blocks is described in Algorithm \ref{algo: counterexample}. The reason for controlling the initial condition for subsequent falsification is to obtain good coverage and diversity in the falsifying traces. Note that we applied slight modifications to this routine. For clarity, these were omitted in the presentation of Algorithm \ref{algo: counterexample} - details are discussed in Section \ref{results}. 

\begin{algorithm}[ht]
	\KwData{Traces of driver behavior: $X^+$; State space discretization: $\mathcal{X}_0$; PSTL formulae: $\Phi$; Dynamical model: $\Sigma$}
	\KwResult{Falsifying traces: $X^{-}=\{(x_0,u_0,...,,x_{N},u_N):\exists \phi \in \Phi, \exists \mathbf{p} \in \mathcal{P}(\phi), \exists x_0\in\mathcal{X}_0, $ such that $\xi_{u_0,...,u_{N-1}}(x_0) \not\models \phi(\mathbf{p})\}$}
	$\mathcal{P} \gets \emptyset$;	$X^{-}\gets\emptyset$; \\
	\ForAll{$\phi \in \Phi$}
	{
		\tcp{Parameter synthesis \cite{asarin2011}}
		$\mathcal{P}(\phi) = \textsc{FindParam}(X^+,\phi)$\\
		\ForAll{$\mathbf{p}\in \mathcal{P}(\phi)$} 
		{
			\tcp{Falsification \cite{jin2015}}
			\ForAll{$x_0 \in \mathcal{X}_0$}
			{
				$X' = \textsc{FalsifyAlgo}(\Sigma(x_0),\phi(\mathbf{p}))$\\
				$X^- \gets X^- \cup X'$
			}
		}
	}
	\caption{\textsc{Counterexample Generation}}
	\label{algo: counterexample}
\end{algorithm}

\section{Results and Discussion}\label{results}
\subsection{Driver Behavior as PSTL Formulae}
Following the framework of Section \ref{solution_approach}, we construct a collection of specifications in PSTL to represent common driving norms. A simple example of this is a specification on speed limit: ``never exceed speed limit". While this technically represents a traffic rule, a higher priority traffic norm is to travel with the flow of traffic; therefore, human drivers typically exceed the posted speed limit by some margin. Consequently, we synthesize a PSTL specification based on the traffic rule and parameterize the true speed limit to accommodate following traffic norms:
\begin{equation}\label{eq:Speedspec}
    \phi_{v_{limit}}(\nu) = \square_{[0,T]} (v_x<\nu).
\end{equation}

Aside from rules such as (\ref{eq:Speedspec}), we investigate how driver behaviors vary based on the state of the traffic light. Essentially, this translates into modeling the driver as a switched system where we seek to learn the behavior rules in each traffic light state. Herein, we formulate a PSTL formula for each traffic light state based on a basic traffic rule activated by that particular traffic light state. In (\ref{eq:Speedspec}) and in subsequent specifications, $T$ is the length of the human driver trajectory.

At a green light, a vehicle should move fast enough to avoid blocking traffic:
\begin{equation}\label{eq:Gspec}
\begin{array}{c}
\phi_{G}(\delta,\tau,\nu) = \square_{[0,T]} (((s_{TL}(t)=G) \\
\wedge (d_x(t)>\delta) \wedge (t_{el}(t)>\tau)) \rightarrow (v_x(t)> \nu))
\end{array}
\end{equation}
\emph{Intuition:} If the traffic light has been green ``for some time", and one is ``sufficiently far" from the intersection, then one should ``not drive too slowly". All expressions in quotation marks are represented as parameters in the PSTL formula.

At a yellow light, vehicles may decide to pass or stop:
\begin{equation}\label{eq:Yspec}
\begin{array}{c}
\phi_{Y}(\delta,\nu_0,\nu) = \square_{[0,T]} ((s_{TL}(t)=Y)\wedge (d_x(t)>\delta) \\
\wedge (v_x(0) > \nu_0) \wedge (t_{el}(t)>0.5) \rightarrow \square_{[0,3]} (v_x(t)> \nu))
\end{array}
\end{equation}
\emph{Intuition:} Based on the vehicle speed and distance to the intersection, if one ``recognizes" a yellow light, then one must decide to pass or stop. In reality, the decision is determined by whether the driver perceives $d_x$ to be larger than her accepted/anticipated stopping distance at current speed. If so, the driver will stop.

At a red light, a vehicle should never cross the intersection:
\begin{equation}\label{eq:Rspec}
\begin{array}{c}
\phi_{R}(\delta,\tau,\nu) = \square_{[0,T]} ((s_{TL}(t)=R)\wedge (d_x(t)>\delta) \\
\wedge (t_{el}(t)>\tau) 
\xrightarrow{}(v_x(t) < \nu))
\end{array}
\end{equation}
\emph{Intuition:} If the traffic light has been red ``for some time" and one is ``close" to the intersection, then one should ``drive slowly".

\subsection{Parameter Synthesis Results}
The parameter synthesis module of \textsc{Breach} was applied to find the feasibility domain for (\ref{eq:Speedspec}), (\ref{eq:Gspec}), (\ref{eq:Yspec}), and (\ref{eq:Rspec}). In order to exploit \textsc{Breach}'s binary-search solver for monotonic specifications, we implement an alternation scheme for PSTL formulae with multiple parameters; we found the results to be consistent regardless of the order of alternation.

For the speed limit specification (\ref{eq:Speedspec}), the result of parameter synthesis found the feasible parameter set to be all speeds less than 25.5 m/s. Observe that this is about $60 \%$ over the posted speed limit of 15.6 m/s (35 mph).

\begin{figure}[!ht]
	\begin{subfigure}{\linewidth}
		\includegraphics[width = \linewidth]{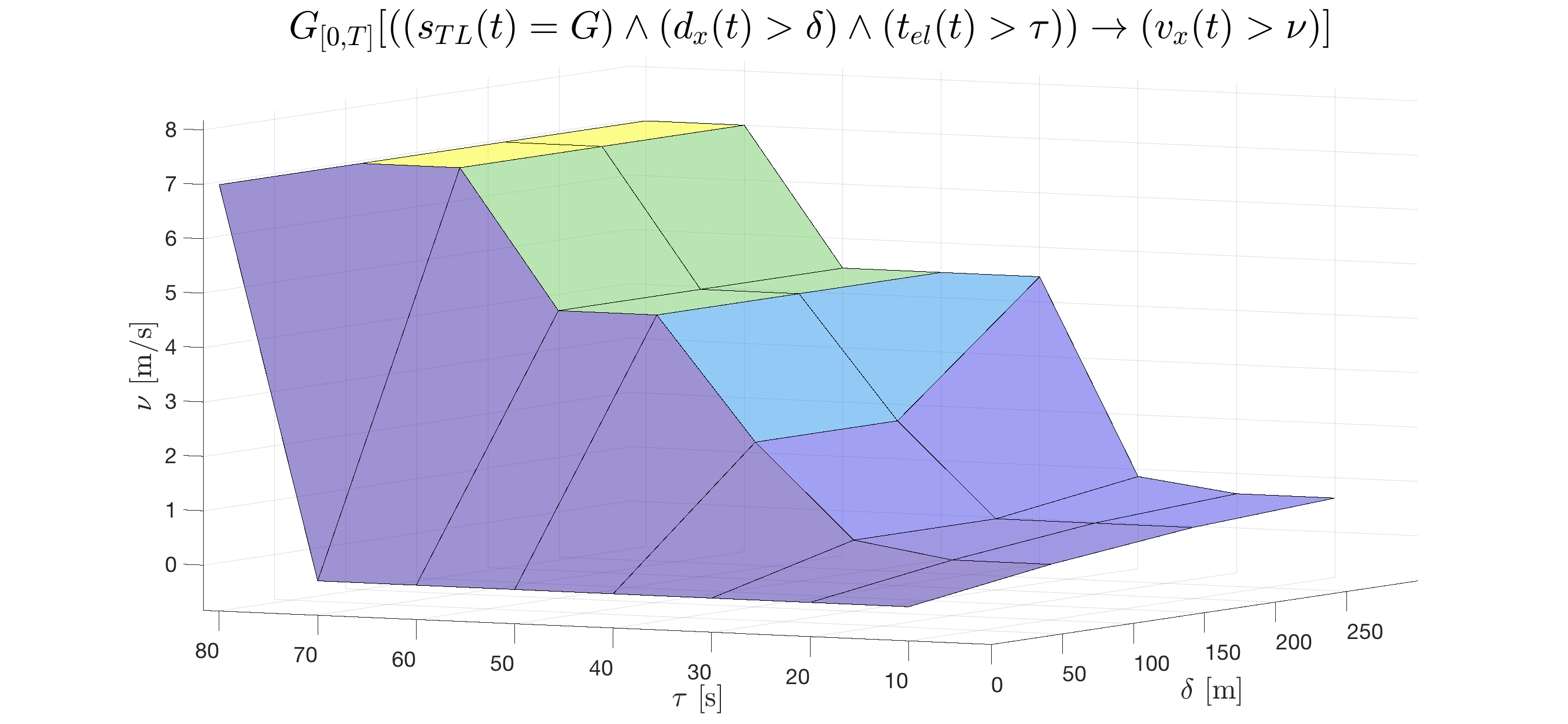}
		\caption{Green light specification: As $\delta$ increases, i.e. the HV is farther from the intersection, the lower bound on velocity, $\nu$, also increases; therefore, if the HV is far away from the intersection, it should drive fast. Furthermore, as $\tau$, the lower bound on $t_{el}$, increases, $\nu$ also increases; therefore, after the traffic light turns green for some time, all the through traffic should not move too slowly.}
		\label{fig:Green}
	\end{subfigure}
	\par\bigskip
	\begin{subfigure}{\linewidth}
		\includegraphics[width = \linewidth]{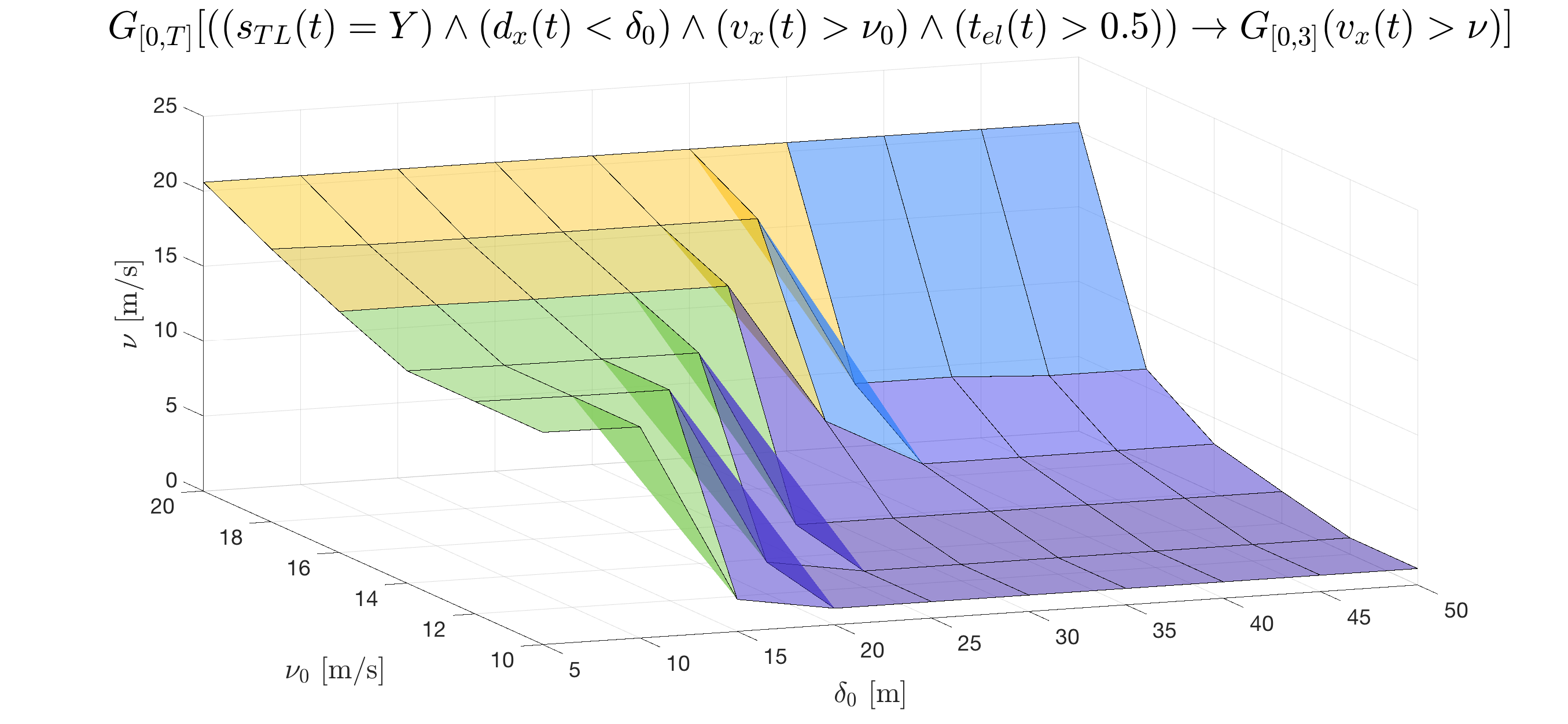}
		\caption{Yellow light specification: If the HV is near the intersection, i.e. $\delta_0$ is small, and it is traveling with a high speed, i.e. $\nu_0$ is large, when the traffic light turns yellow and it is had enough time to register this change, i.e. $t_{el} > 0.5$, then in the following three seconds, the HV will try to pass, i.e. its speed will never drop below a high value of $\nu$. On the other hand, when the vehicle is far away and traveling slowly, i.e. $\delta_0$ is large and $\nu_0$ is small, then it will decelerate in anticipation of the impending red light. Interestingly, we can observe the transition from where $\nu$ changes from a high to low value as a function of the distance to intersection and speed at the time when the HV registered the light change. These results are fairly intuitive.}
		\label{fig:Yellow}
	\end{subfigure}
    \par\bigskip
	\begin{subfigure}{\linewidth}
		\includegraphics[width = \linewidth]{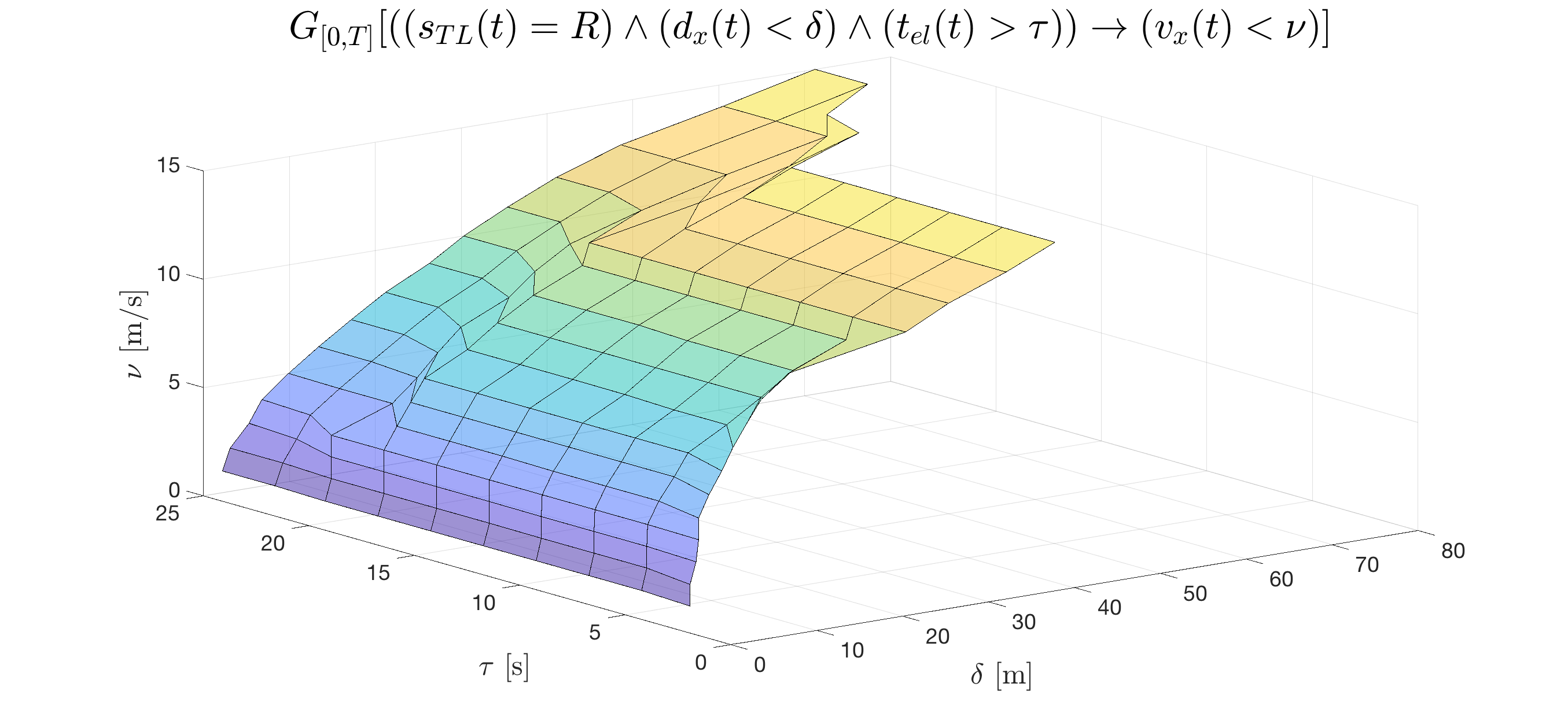}
		\caption{Red light specification: As $\delta$ decreases, the upper bound on velocity, $\nu$ decreases, meaning vehicles tend to slow down when close to the intersection; This trend is similar across all $\tau$.}
		\label{fig:Red}
	\end{subfigure}

	\caption{Validity frontiers for traffic light specifications}
	\label{fig:Case1}
\end{figure}

For the remaining specifications, we found a multi-dimensional Pareto frontier to represent the boundary of the feasible parameter set as in \cite{jin2015}, \cite{asarin2011}. These frontiers are illustrated in Figure \ref{fig:Case1}.

\subsection{Falsification Results}
The falsification routine as described in Section \ref{section: falsification} requires a parameter set, $\mathcal{P}$; hence we sample parameters from the Pareto frontier in Figure \ref{fig:Case1} in order to ensure we obtain good coverage and diversity in the falsifying traces.

The falsification problem requires classes of input signals, which are used together with a dynamical model to create falsifying traces. In this work, we consider piece-wise constant signals of duration 0.5 seconds; during each constant segment, the input can assume a value within the set $U = [-6, 3] $ m/s$^2$. The simulation horizon is three seconds and thus the input signal contains six control points.

\begin{figure}[ht]
	\begin{subfigure}[b]{0.48 \linewidth}
		\includegraphics[width = \linewidth]{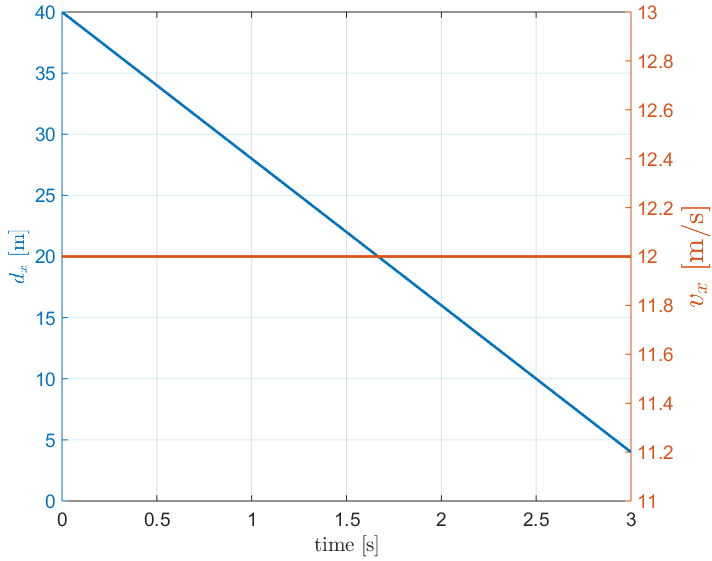}
		\caption{A falsifying trace of the red light specification}
		\label{fig:Trace}		
	\end{subfigure}
	\begin{subfigure}[b]{0.48 \linewidth}
		\includegraphics[width = \linewidth]{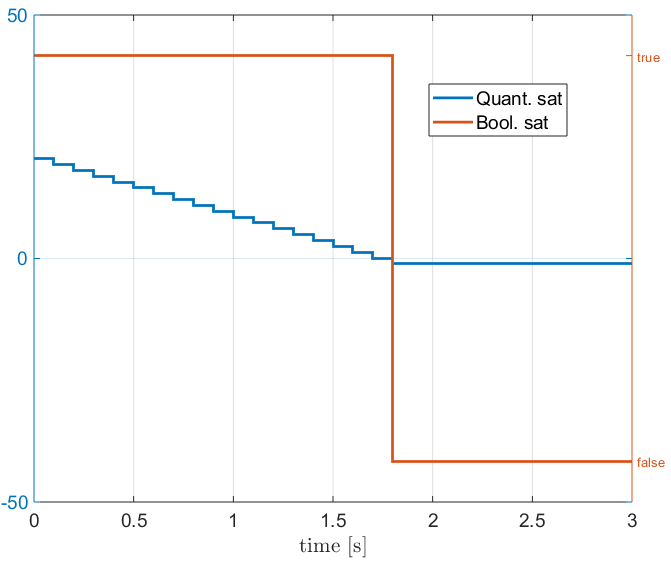}
		\caption{Robust satisfaction along the trace}
		\label{fig:Sat}		
	\end{subfigure}
	\caption{An example of falsification of the STL formula $\phi_{R}(19.5,7.5,10)$}
	\label{fig:False}
\end{figure}

Figure \ref{fig:Trace} illustrates a falsifying trajectory of the red light specification 
\[
\begin{array}{c}
\phi_{R}(19.5,7.5,10) = \square_{[0,T]} ((s_{TL}(t)=R)\wedge (d_x(t)<19.5) \\
\wedge (t_{el}(t)>7.5) \rightarrow (v_x(t) < 10)).
\end{array}
\]
Here, the HV simply maintains its speed when approaching the intersection, and thus violates the spec on ``HV should lower its speed as it approaches the intersection". Figure \ref{fig:Sat} shows the robust satisfaction of 
\[
(s_{TL}(t)=R)\wedge (d_x(t)<19.5) \wedge (t_{el}(t)>7.5)\rightarrow (v_x(t) < 10)
\]
which is the portion of $\phi_{R}(19.5,7.5,10)$ within the ``always".

Observe that the violation given in Figure \ref{fig:False} is almost trivial. To avoid only generating such trivially falsifying traces, we use the following strategy: (1) Use CMA-ES solver instead of the Nelder-Mead method; (2) Apply a difference metric criterion to select diverse falsifying traces; (3) Accept traces with sub-optimal robustness violation.

Experimental results showed that the CMA-ES solver produced more diverse input signals resulting in specification violation than the simplex-based Nelder-Mead method \cite{press2007}. The difference criterion involved checking that the Euclidean distance between two candidate falsifying inputs was sufficiently large to avoid repetition of the same signals. And finally, accepting sub-optimal robustness violations allowed for generating counterexamples closer to the expected boundary between ``human" and ``non-human" behavior.

Using this strategy together with the falsification method described previously, we found 170 falsifying traces for (\ref{eq:Speedspec}); 7,068 falsifying traces for (\ref{eq:Gspec}); 21,784 falsifying traces for (\ref{eq:Yspec}); and 2,926 falsifying traces for (\ref{eq:Rspec}). Note that additional falsifying traces can be generated by increasing the maximum iterations allowed for the solver.

\subsection{Classification Results}
In our initial treatment of the classification problem, we construct individual classifiers for each state of the traffic light to address the hybrid nature of this system. Furthermore, we take measures to make the classification task more computationally tractable by sub-sampling the training examples and omitting the queue length ($l_q$). Elimination of the queue length from this initial analysis is justified for the green traffic light state because the queue length is always zero; additionally, our specification for the red traffic light does not incorporate the queue length and consequently, the queue length does not factor into deciding whether or not a trace falsifies or satisfies the specification. Sub-sampling the traces from 10 Hz to 2 Hz reduces the trace sizes by 5 times. The number of sub-sampled traces (both positive and negative traces) are roughly 11,000 for $s_{TL} = G$, and 6,000 for $s_{TL}=R$. Finally, 70\% and 30\% of the sub-sampled traces were each used as the training and the testing sets.

An overview of the individual classifiers is given in the following:
\begin{itemize}
    \item \emph{Green traffic light classification:} negative examples are traces found to violate (\ref{eq:Gspec}) and the features considered are $X(t) = [d_x(t), v_x(t), t_{el}(t), u(t)]$. For MLP classifiers, the input to the classifier is a flattened sequence, $\xi(t) = [X(t-3.0), X(t-2.5), ..., X(t)]$ and the input to the RNN classifier is the sequence of $X(t)$.
    \item \emph{Red traffic light classification:} negative examples are traces found to violate (\ref{eq:Rspec}). Since $l_{q}$ is not considered in the spec, we omit it from this analysis. The resulting features are $X(t) = [d(t), v(t), T_{el}(t), a(t)]$. Inputs to the MLP and RNN classifiers are the same as those of the green classifier.
\end{itemize}
Herein, we have only constructed classifiers for the green and the red traffic lights, leaving treatment of the yellow light as future work for the following reason: the duration of the yellow light is short, and transitions between green/yellow and yellow/red are important and can indeed take place during the considered horizon. However, currently the negative training example are generated with a constant traffic light state and thus the transitions themselves are not well captured; furthermore, there are very few positive training examples for a yellow traffic light.

The MLP is modeled with a dense layer with 28 hidden units, ReLu activation, and soft-max function at the end of the network. The RNN is modeled with a recurrent layer containing 36 hidden units, ReLu activation, and soft-max function at the end of the network. We used categorical cross entropy as our loss function, and the ADAM optimizer. 

Table \ref{tab:comparison} summarizes the (converged) accuracy of the two classifiers when tested on a test set. 
\begin{table}[ht]
    \centering
    \begin{tabular}{cc|cc} 
    \multicolumn{2}{c|}{$s_{TL} = G$} & \multicolumn{2}{c}{$s_{TL} = R$} \\
    \hline
    MLP & RNN & MLP & RNN  \\ \hline
    99.4 & 99.8 & 99.7 & 99.9 \\ \hline
    \end{tabular}
    \caption{Comparison of MLP and RNN for different traffic light states}
    \label{tab:comparison}
\end{table}

We speculate that one reason for the extremely high accuracy is that the classification was too easy or trivial for much of the data. This indicates that perhaps the falsified trajectories were too far away from the true boundary between the ``human" and ``non-human" classes. Possible rectifications to this issue are addressed in Section \ref{conclusion}.

\begin{figure}[ht]
	\begin{subfigure}[b]{0.48\linewidth}
		\includegraphics[width=\linewidth]{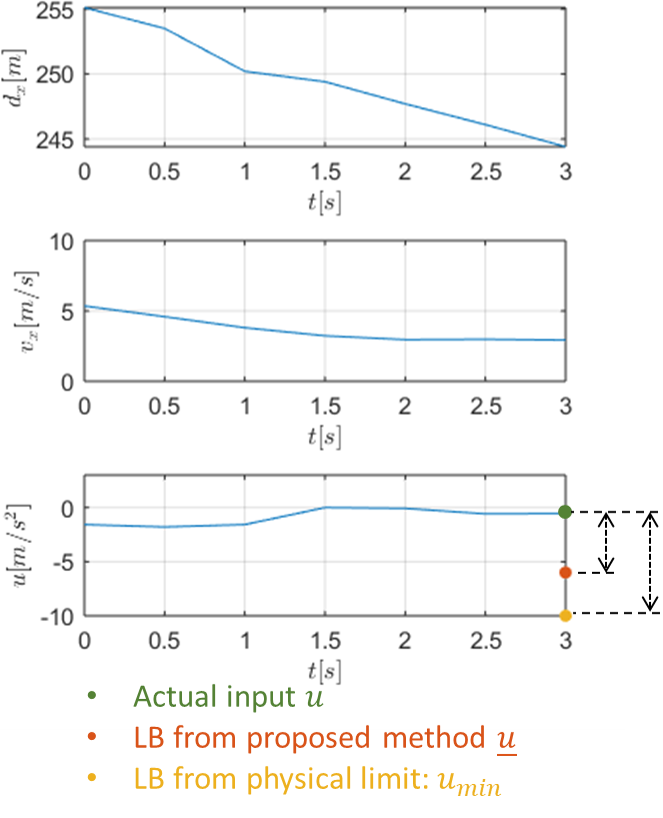}
		\caption{}
		\label{fig:BD03}
	\end{subfigure}
	\begin{subfigure}[b]{0.48\linewidth}
		\includegraphics[width=\linewidth]{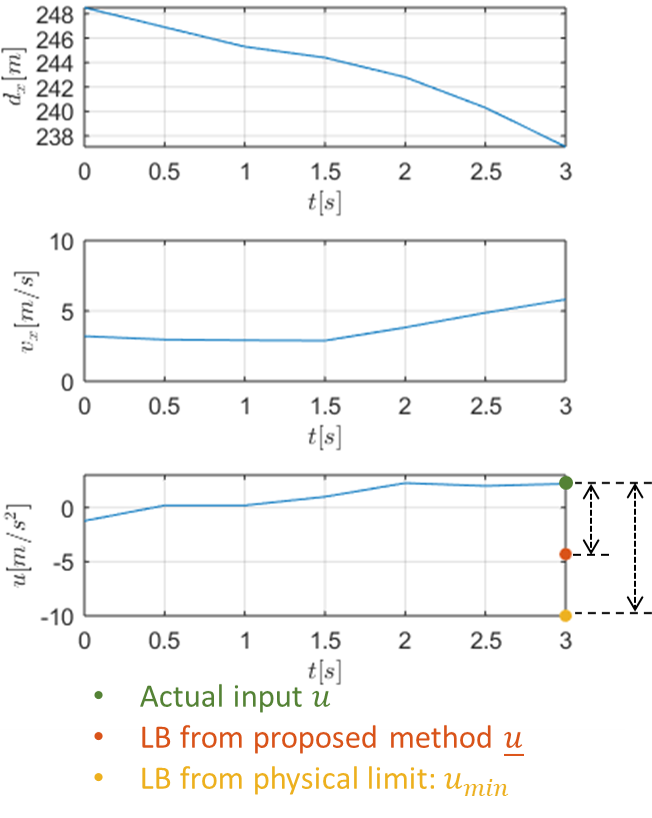}
		\caption{}
		\label{fig:BD04}
	\end{subfigure}
	\caption{Examples of the resulting bound on HV acceleration}
\end{figure}

 We examine the generated bound on HV acceleration for some cases where the classifier was effective at reducing uncertainty. Next, we briefly describe the querying process for computing the the set of ``human" accelerations given a classifier. For a vector of states and inputs in the horizon,
 \[
 \xi_0 = [x(0),x(1),...x(T),u(0),u(1)...u(T-1)]^T,
 \] 
 we sweep the next input signal across the entire range of acceleration to form $\{u_i(T)\},\text{where }i=1,2,3...$; each $u_i(T)$ is used to complete a vector 
 \[
 \xi_0^i = [\xi_0^T, u_i(T)]^T.
 \]
 Next, the $\xi_0^i$ are passed into the classifier. Finally, ``human" inputs are defined to be:
 \[
 \{u_i(T) ~|~ p_H(\xi_0^i) \geq 0.5\}.
 \]
 
 In Figures \ref{fig:BD03} and \ref{fig:BD04}, we plot two 3-second traces extracted from human naturalistic driving data. For each trace, the lower bound on next acceleration input $u(T)$ is estimated using the querying routine above. The yellow circle marks the estimated input based on the physical acceleration limit of the HV, which is always $-10$ m/s$^2$; the red circle marks the estimated lower bound of acceleration from the classification method, while the green circle shows the actual acceleration undertaken by the HV. The proposed method shows a tighter acceleration bound in comparison to physical limits in both cases while remaining below the actual acceleration, i.e. being conservative. In case (a), the HV is already moving at a low speed, so it is intuitive that it will not suddenly conduct full brake; in case (b), the HV accelerates from a low speed, so it is unlikely to suddenly brake at the next instant. However, we also observed that the bound from the proposed method is still very conservative, as braking with acceleration around $-5$ m/s$^2$ can be already perceived as hard brake; moreover, the method does not always give tighter bound than the physical limit.
 
 \section{Conclusion and Future Work}\label{conclusion}
In this work, we proposed a framework to construct a data-driven bound on human driver behavior that allows for verifying whether a given trajectory originates from a human driver. Our results and contributions are summarized below:
\begin{itemize}
    \item Generation of data-driven bounds on HV acceleration. From the perspective of control synthesis, the benefit of tighter bounds on human action is less uncertainty about the disturbance.
    \item Synthesis of reasonable specifications for HV behavior.
    \item Generation of ``non-human behavior" as falsifying traces of STL formulae.
    \item Construction of classifiers to distinguish between human and non-human driving traces.
\end{itemize}

This work is a first step in using falsification-based generation of negative training examples. Consequently, many avenues should be explored to improve the performance of the proposed framework. In particular, the classifier gave useful results for some traces, but failed to restrict the bound on human acceleration for many others. This is likely due to the high dimensionality of the problem, since our approach seeks to leverage information over a time horizon. Furthermore, only a subsets of the 556 trajectories were considered for a given specification. Consequently, the training set may be insufficient. Thus possible future approaches may include seeking a larger data set or shortening the time horizon to reduce problem dimensionality. 

Additionally, negative training examples were generated by considering piece-wise constant input signals, which often featured large differences between constant segments. For instance, an input signal could be constant at $-6$ m/s$^2$ during the $[0,0.5)$ s interval before changing to $3$ m/s$^2$ during the $[0.5,1)$ s interval. However, HV accelerations do not feature such excursions. Consequently, it is possible that many of the generated negative training examples were very far from positive training examples in the feature space of the classification problem. Therefore, a well-performing classifier may indeed find the boundary to be very close to the negative training examples and as a result, deem many actions that intuitively appear to be non-human as human. A potential remedy that will be considered in future work is attempt falsification using a class of smooth input signals and again accept traces with sub-optimal robustness violations. If the resulting negative training examples are closer to the positive training examples in the feature space, then we can expect a well-performing classifier to be more discerning between human and non-human behavior. To address the risk of potentially over-fitting to the observed data, the iterative method introduced briefly in Section \ref{solution_approach} may be of value. The core idea is to follow the procedure of Section \ref{solution_approach} to generate a nominal classifier, and then sample this classifier near its boundary points to augment the set of positive training examples, $X^+$, before repeating the procedure of Section \ref{solution_approach}. By augmenting $X^+$, we speculate that the parameter synthesis method will find a larger feasible parameter domain and consequently ``push out" the classifier towards more negative training examples.

In addition to reducing uncertainty by determining tighter bounds on human action, it is also important to have a notion of uncertainty quantification. A classifier based upon convex programming principles can offer this quality through the notion of an upper limit on the probability of a new observation violating the constructed input bound \cite{chen2018}, \cite{calafiore2009}. However, the approach of \cite{chen2018} considered stationary points as opposed to the time series considered here, which add additional complexity.

Finally, re-visiting the motivation of this work, we believe our framework can be applied to the synthesis of safe \& optimal controllers and the identification of corner cases for controller evaluation. The critical ingredient in achieving this objective will be computing reachable sets using the state-dependent disturbance bounds induced by our approach.

%%%%%%%%%%%%%%%%%%%%%%%%%%%%%%%%%%%%%%%%%%%%
\section*{Acknowledgment}
The authors would like to thank Prof. Necmiye Ozay and Dr. Alex Donz\'e for their valuable insights and instructive conversations.
%%%%%%%%%%%%%%%%%%%%%%%%%%%%%%%%%%%%%%%%%%%%

\bibliographystyle{IEEEtran}
\bibliography{references}

\end{document}